\documentclass[11pt,a4paper]{article}
\usepackage[hyperref]{acl2021}
\usepackage{times}
\usepackage{latexsym}

\usepackage{mathtools}
\newcommand\Set[2]{\{\,#1\mid#2\,\}}

\usepackage{xurl}
\usepackage{subcaption}
\usepackage{amsmath}
\usepackage{multirow}
\usepackage{makecell}
\usepackage{booktabs}
\usepackage{bm}
\usepackage{bbm}
\usepackage{amssymb}

\usepackage{graphicx} 
\usepackage{float} 
\usepackage[noend]{algpseudocode}
\usepackage{algorithmicx,algorithm}
\usepackage{microtype}

\makeatletter
\newcommand{\printfnsymbol}[1]{%
  \textsuperscript{\@fnsymbol{#1}}%
}
\makeatother

\aclfinalcopy 

\title{Selective Knowledge Distillation for Neural Machine Translation}

\author{Fusheng Wang\thanks{~ Equal contribution.}\,\,\thanks{~ This work was done when Fusheng Wang was interning at Pattern Recognition Center, Wechat AI, Tencent Inc, China.}$\,\,^{1}$, Jianhao Yan$^{* 2}$ , Fandong Meng$^{2}$ ,  Jie Zhou$^{2}$ \\
Peking University, China$^{1}$ \\
Pattern Recognition Center, WeChat AI, Tencent, China$^{2}$ \\
{wfs0315@pku.edu.com} \\
{\{elliottyan, fandongmeng\}@tencent.com} \\
}

\date{}

\begin{document}
\maketitle
\begin{abstract}
Neural Machine Translation (NMT) models achieve state-of-the-art performance on many translation benchmarks. 
As an active research field in NMT, knowledge distillation is widely applied to enhance the model's performance by transferring teacher model's knowledge on each training sample.
However, previous work rarely discusses the different impacts and connections among these samples, which serve as the medium for transferring teacher knowledge.
In this paper, we design a novel protocol that can effectively analyze the different impacts of samples by comparing various samples' partitions.
Based on above protocol, we conduct extensive experiments and find that the teacher's knowledge is not the more, the better.
Knowledge over specific samples may even hurt the whole performance of knowledge distillation. 
Finally, to address these issues, we propose two simple yet effective strategies, i.e., batch-level and global-level selections, to pick suitable samples for distillation. 
We evaluate our approaches on two large-scale machine translation tasks, WMT'14 English-German and WMT'19 Chinese-English. Experimental results show that our approaches yield up to +1.28 and +0.89 BLEU points improvements over the Transformer baseline, respectively. \footnote{We release our code on \url{https://github.com/LeslieOverfitting/selective_distillation}.}
\end{abstract}

\section{Introduction}
Machine translation has made great progress recently by using sequence-to-sequence models \citep{sutskever2014sequence, vaswani2017attention,meng2019dtmt,zhangetal2019bridging,yanetal2020multi}.
Recently, some knowledge distillation methods \citep{kim2016sequence, freitag2017ensemble, gu2017non, tan2019multilingual, Wei2019OnlineDF, li2020learning, wu2020skip} are proposed in the machine translation to help improve model performance by transferring knowledge from a teacher model. 
These methods can be divided into two categories: word-level and sequence-level, by the granularity of teacher information.
In their researches, the model learns from teacher models by minimizing gaps between their outputs on every training word/sentence (i.e., corresponding training sample) without distinction.

Despite their promising results, previous studies mainly focus on finding what to teach and rarely investigate how these words/sentences (i.e., samples), which serve as the medium or carrier for transferring teacher knowledge, participate in the knowledge distillation.
Several questions remain unsolved for these samples:
Which part of all samples shows more impact in knowledge distillation? 
Intuitively, we may regard that longer sentences are hard to translate and might carry more teacher knowledge.
But are there more of these criteria that can identify these more important/suitable samples for distillation? 
Further, what are the connections among these samples? Are they all guiding the student model to the same direction?
By investigating the carrier of teacher knowledge, we can shed light on finding the most effective KD method.

Hence, in this paper, we aim to investigate the impacts and differences among all samples. 
However, it is non-trivial to analyze each of them. 
    Therefore, we propose a novel analytical protocol by partitioning the samples into two halves with a specific criterion (e.g., sentence length or word cross-entropy) and study the gap between performance.
Extensive empirical experiments are conducted to analyze the most suitable sample for transferring knowledge.
We find that different samples differ in transferring knowledge for a substantial margin.
More interestingly, with some partitions, especially the student model's word cross-entropy, the model with half of the knowledge even shows better performance than the model using all distill knowledge.
The benefit of the distillation of two halves cannot collaborate.
This phenomenon reveals that the distillation of two halves cannot collaborate, even hurt the whole performance. Hence, a more sophisticated selective strategy is necessary for KD methods. 

Next, we propose two simple yet effective methods to address the observed phenomenon according to word cross-entropy (Word CE), which we find is the most distinguishable criterion. 
We first propose a batch-level selection strategy that chooses words with higher Word CE within the current batch's distribution. 
Further, to step forward from local (batch) distribution to global distribution, we use a global-level FIFO queue to approximate the optimal global selection strategy, which caches the Word CE distributions across several steps.
We evaluate our proposed method on two large-scale machine translation datasets: WMT'14 English-German and WMT'19 Chinese-English. Experimental results show that our approach yields an improvement of +1.28 and + 0.89 BLEU points over the Transformer baseline.

In summary, our contributions are as follows:
\begin{itemize}
\item We propose a novel protocol for analyzing the property for the suitable medium samples for transferring teacher's knowledge.
\item We conduct extensive analyses and find that some of the teacher's knowledge will hurt the whole effect of knowledge distillation.
\item We propose two selective strategies: batch-level selection and global-level selection. The experimental results validate our methods are effective.
\end{itemize}

\section{Related Work}
Knowledge distillation approach~\citep{hinton2015distilling} aims to transfer knowledge from teacher model to student model. Recently, many knowledge distillation methods~\citep{kim2016sequence,hu2018attention,sun2019patient,tang2019distilling,jiao2019tinybert,zhang2019future,  zhang2020ternarybert,chenetal2020bridging,meng-etal-2020-wechat} have been used to get effective student model in the field of natural language processing by using teacher model's outputs or hidden states as knowledge. 

As for neural machine translation (NMT), knowledge distillation methods commonly focus on better improving the student model and learning from the teacher model.
\citet{kim2016sequence} first applied knowledge distillation to NMT and proposed the sequence-level knowledge distillation that lets student model mimic the sequence distribution generated by the teacher model. It was explained as a kind of data augmentation and regularization by \citet{gordon2019explaining}. Further, \citet{freitag2017ensemble} improved the quality of distillation information by using an ensemble model as the teacher model. \citet{gu2017non} improved non-autoregressive model performance by learning distillation information from the autoregressive model. \citet{wu2020skip} proposed a layer-wise distillation method to be suitable for the deep neural network. \citet{chen2020distilling} let translation model learn from language model to help the generation of machine translation.

To the best of our knowledge, there is no previous work in NMT concerning the selection of suitable samples for distillation.
The few related ones mainly focus on selecting appropriate teachers for the student model to learn. 
For instance, ~\citet{tan2019multilingual} let the student model only learn from the individual teacher model whose performance surpasses it.
~\citet{Wei2019OnlineDF} proposed an online knowledge distillation method that let the model selectively learn from history checkpoints. 
Unlike the above approaches, we explore the effective selective distillation strategy from sample perspective and let each sample determine learning content and degree.

\section{Background}
\subsection{Neural Machine Translation}
Given a source sentence $\bm{x}=(x_1,...,x_n)$, and its corresponding ground-truth translation sentence $\bm{y}=(y_1^*,...,y_m^*)$, an NMT model minimizes the word negative log-likelihood loss at each position by computing cross-entropy. For the $j$-th word in the target sentence, the loss can be formulated as:
{
\setlength\abovedisplayskip{1.5pt}
\setlength\belowdisplayskip{1.5pt}
\begin{align} \label{cross_entorpy}
    \hspace{-2mm}
    \mathcal{L}_{ce}  = - \sum_{k=1}^{|V|}{\mathbbm{1}\{y_j^* = k\} \log p(y_j = k| \bm{y}_{<j},\bm{x}; \theta)},
\end{align}
}
where $|V|$ is the size of target vocabulary, $\mathbbm{1} $ is the indicator function, and $p(\cdot |\cdot)$ denotes conditional probability with model parameterized by $\theta$.
\subsection{Word-level Knowledge Distillation}
\label{word_kd}
In knowledge distillation, student model $S$ gets extra supervision signal by matching its own outputs to the probability outputs of teacher model $T$. Specifically, word-level knowledge distillation defines the Kullback–Leibler distance between the output distributions of student and teacher~\cite{hu2018attention}. 
After removing constants, the objective is formulated as:
{
\setlength\abovedisplayskip{0.5pt}
\setlength\belowdisplayskip{0.5pt}
\begin{align} \label{kd}
    \mathcal{L}_{kd} =  - \sum_{k=1}^{|V|} q(y_j=k| \bm{y}_{<j},\bm{x};\theta_{T}) \notag  \\
    \times \log p(y_j=k| \bm{y}_{<j},\bm{x}; \theta_{S}) ,
\end{align}
}where $q(\cdot |\cdot)$ is the conditional probability of teacher model. $\theta_{S}$ and $\theta_{T}$ is the parameter set of student model and teacher model, respectively.

And then, the overall training procedure is minimizing the summation of two objectives:
\begin{align} \label{total_object}
\mathcal{L} = \mathcal{L}_{ce} + \alpha \mathcal{L}_{kd},
\end{align} 
where $\alpha$ is a weight to balance two losses.

\section{Are All Words Equally Suitable for KD?}
\label{sec:sec3}
\begin{table}[!t]
\begin{center}
\begin{tabular}{|cc|c|c|c|}
 \hline
 \multicolumn{2}{|l|}{\multirow{2}{*}{Criteria}} &
 \multicolumn{3}{c|}{BLEU} \\
 \cline{3-5}
  \multicolumn{2}{|c|}{} & $\mathcal{S}_{High}$ & $\mathcal{S}_{Low}$ & $\Delta$ \\
 \hline
  \multicolumn{2}{|l|} {Baseline} &  \multicolumn{2}{c|}{27.29 } & {-} \\ \hline
     \multicolumn{2}{|l|} {Distill-All} &  \multicolumn{2}{c|}{28.14 } & {-} \\ \hline
 \multicolumn{2}{|l|} {Distill-Half(Random) } &  \multicolumn{2}{c|}{28.18 } & {-} \\ \hline
 \multicolumn{5}{|c|} {Data Property} \\ \hline
 \multicolumn{2}{|l|}{Sentence Length} & 27.81 & 27.59 & +0.22 \\
 \multicolumn{2}{|l|}{Word Frequency} & 28.35 & 27.99 & +0.36* \\ \hline
  \multicolumn{5}{|c|} {Student Model} \\ \hline
 \multicolumn{2}{|l|}{Embedding Norm} & 27.90 & 27.73 & +0.17 \\
 \multicolumn{2}{|l|}{Word CE} & \textbf{28.42} & 27.78 & \textbf{+0.64*} \\
 \multicolumn{2}{|l|}{Sentence CE} & 28.29 & 27.84 & +0.45* \\
 \hline
  \multicolumn{5}{|c|} {Teacher Model} \\ \hline
 \multicolumn{2}{|l|}{Teacher $P_{golden}$} & 27.97 & 28.00 & -0.03 \\
 \multicolumn{2}{|l|}{Entropy} & 27.62 & 27.92 & -0.30 \\
 \hline
 \end{tabular}
\caption {BLEU score (\%) of different criteria in WMT'14 En-De.
$\Delta$ denotes the difference of BLEU score (\%) between $\mathcal{S}_{High}$ and $\mathcal{S}_{Low}$. `*': significantly ($p < 0.05$) difference between the $\mathcal{S}_{High}$ and $\mathcal{S}_{Low}$ .}
\label{tab:performance_criteria}
\end{center}
\end{table}

As discussed before, as a carrier of the teacher's knowledge, ground-truth words might greatly influence the performance of knowledge distillation.
Therefore, in this section, we first do some preliminary empirical studies to evaluate the importance of different words/sentences in knowledge distillation.

\subsection{Partition of Different Parts}
The optimal way to analyze samples' different impacts on distillation is to do ablation studies over each of them. However, it is clearly time-consuming and intractable. 
Hence, we propose an analytical protocol by using the partition and comparison as an approximation, which we believe could shed light on future analyses.
Particularly, we leverage a specific criterion $f$ to partition samples into two complementary parts:
\begin{align*}
    \mathcal{S}_{High}  &\coloneqq \Set{y_i}{ \operatorname{f}(y_i) > \operatorname{Median}(f(\bm{y})), y_i\in \bm{y}},\\
    \mathcal{S}_{Low}  &\coloneqq \Set{y_i}{\operatorname{f}(y_i)\leq \operatorname{Median}(f(\bm{y})), y_i\in \bm{y}},
\end{align*}
and analyze different effects between $\mathcal{S}_{High}$ and $\mathcal{S}_{Low}$.
Each part consists of 50\% words/sentences precisely. 
The criteria come from three different perspectives: data property, student model, and teacher model. The detailed descriptions are as follows:

\begin{itemize}
\vspace{-10pt}
\setlength{\itemsep}{3pt}
\setlength{\parsep}{0pt}
\setlength{\parskip}{0pt}

\item \textbf{Data Property.} 

As longer sentences and rare words are more challenging to translate \cite{kocmi2017curriculum,platanios2019competence}, its
corresponding teacher knowledge may benefit the student model more. Hence, we choose sentence length and word frequency as criteria.

\item \textbf{Student Model.} 
As for the student model, we care if the student model thinks these words/sentences are too complicated.
Therefore, we use Word CE (cross-entropy of words), Sentence CE (mean of the cross-entropy of all words in sentences), and each word's embedding norm \cite{liu2020norm}.

\item \textbf{Teacher Model.} For the teacher model, we guess that the teacher's prediction confidence may be crucial for transferring knowledge. Hence, we use the prediction probability of ground-truth label ($P_{golden}$) and entropy of prediction distribution as our criteria.
\end{itemize}

\subsection{Analytic Results}

Table \ref{tab:performance_criteria} presents our results on different criteria. We also add the performance of Transformer baseline, Distill-All (distillation with all words) and Distill-Half(distillation with 50\% words chosen by random) for comparison. 

\paragraph{Impact of Different Parts.} Through most of the rows, we observe noticeable gaps between the BLEU scores of the $\mathcal{S}_{High}$ and $\mathcal{S}_{Low}$, indicating there exists a clear difference of impact on medium of teacher knowledge. 
Specifically, for most of the criteria like cross-entropies or word frequency, the gap between two halves surpasses 0.35. 
In contrast, teacher $P_{golden}$ seems not useful for partitioning KD knowledge. 
We conjecture this is because no matter whether the teacher is convinced with the golden label or not, other soft labels could contain useful information \cite{gouetal2020knowledge}. 
Besides, we find teacher entropy is a good-enough criterion for partitioning KD data, which inlines with previous studies of dark knowledge \cite{dong2019distillation}.
Finally, we find that the KD is most sensitive (+0.64) with the Word CE criterion, which enjoys the adaptivity during the training phase and is a good representative for whether the student thinks the sample is difficult. 

In conclusion, we regard the most suitable samples should have the following properties: higher Word CE, higher Sentence CE, higher Word Frequency, which probably benefits future studies of effective KD methods.

\paragraph{Impact of All and Halves.} 
More interestingly, compared with `Distill-All', which is the combination of the $\mathcal{S}_{High}$ and $\mathcal{S}_{Low}$, the $\mathcal{S}_{High}$ halves' BLEU score even surpass the `Distill-All', for Word CE, Sentence CE and Word Frequency
criteria.
This leads to two conclusions:

(1) Within some partitions, the $\mathcal{S}_{High}$ contributes most to the KD improvements.

(2) The amount of teacher knowledge is not the more, the better. The distillation knowledge of the $\mathcal{S}_{Low}$ does not directly combine with the $\mathcal{S}_{High}$, even hurts $\mathcal{S}_{High}$'s performance. 

\begin{figure}[!t] 
\centering 
\includegraphics[width=0.45\textwidth]{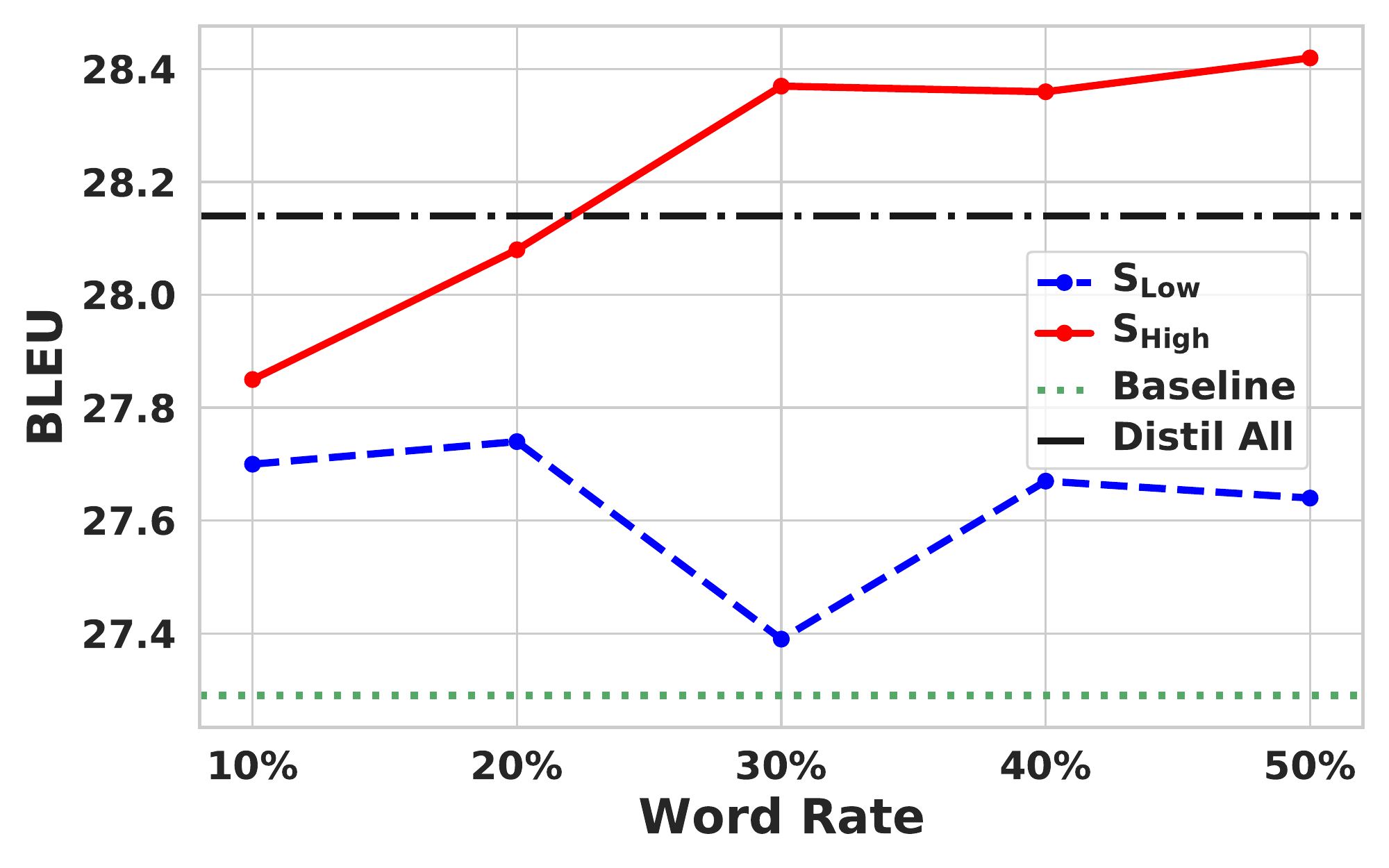} 
\caption{BLEU score (\%) on WMT'14 En-De translation task. $\mathcal{S}_{Low}$ means the subset of training set which have relative small word-level cross-entropy and easy for model to learn. $\mathcal{S}_{High}$ means the subset of training set which have relative large word-level cross-entropy and hard for model to learn. `Word Rate' controls the number of words need to get extra distillation knowledge from teacher model. For example, word rate=30\% means that student model only learns distillation knowledge of words whose cross-entropy loss in biggest / smallest 30\%. We choose the model which performs the best on the validation set and report its performance on test sets.}
\vspace{-15pt}
\label{Fig.word_ce_easy_hard}
\end{figure}

\paragraph{Impact of the Amount of Knowledge.} 
Given that distillation knowledge is most sensitive to Word CE, we conduct extra analysis on the Word CE. 
Figure \ref{Fig.word_ce_easy_hard} presents the results of varying the amount of knowledge for $\mathcal{S}_{High}$ and $\mathcal{S}_{Low}$. 
The consistent phenomenon is that the $\mathcal{S}_{High}$ perform significantly better than the $\mathcal{S}_{Low}$ when using the same amount of teacher's knowledge. 
These results suggest that we should focus more on the $\mathcal{S}_{High}$ than on $\mathcal{S}_{Low}$.
Besides, we notice that the model performance increases when we increase the knowledge in $\mathcal{S}_{High}$, but not the case for $\mathcal{S}_{Low}$. 
We conclude that the Word CE is distinguishable and a better indicator of teachers' useful knowledge only for $\mathcal{S}_{High}$.

At the end of this section, we can summary the following points:
\begin{itemize}

\setlength{\itemsep}{3pt}
\setlength{\parsep}{0pt}
\setlength{\parskip}{0pt}

\item To find out the most suitable medium for transferring medium, we adopt a novel method of partition and comparison, which can easily be adopted to future studies.
\item The benefit of distillation knowledge drastically changes when applying to different mediums of knowledge. 
\item Among all criteria, knowledge distillation is the most sensitive to Word CE.
Distilling words with higher Word CE is more reliable than words with lower CE.
\item In some partitions, the distillation benefit of $\mathcal{S}_{Low}$ can not add to the $\mathcal{S}_{High}$, even hurts $\mathcal{S}_{High}$'s performance.
\end{itemize}

\section{Selective Knowledge Distillation for NMT}
\label{sec:selective_distillation}
As mentioned above, there exist un-suitable mediums/samples that hurt the performance of knowledge distillation. 
In this section, we address this problem by using two simple yet effective strategy of selecting useful samples.

In Section 4, we find that Word CE is the most distinguishable criterion. Hence, we continue to use the Word CE as the measure in our methods. 
As the word cross-entropy is a direct measure of how the student model agrees with the golden label, we refer to words with relatively large cross-entropy as difficult words, and words with relatively small cross-entropy as easy words, in the following parts. This is to keep the notation different from previous analysis. 

Then, we only need to define what is ``relatively large''.
Here, we introduce two CE-based selective strategies:
\vspace{-5pt}
\paragraph{Batch-level Selection (BLS).} 
Given a mini-batch $B$ of sentence pairs with $M$ target words, we sort all words in the current batch with their Word CE in descending order and select the top $r$ percent of all words to distill teacher knowledge. 
More formally, let $\mathcal{A}$ denote the Word CE set, which contains the Word CE of each word in batch B. 
We define $\mathcal{S}_{Hard} = top\_{r\%}(\mathcal{A})$ as the set of the  $r\%$ largest cross-entropy words among the batch, and $\mathcal{S}_{Easy}$ is its complementary part.

For those words in $\mathcal{S}_{Hard}$, we let them get extra supervision signal from teacher model's distillation information. Therefore, the knowledge distillation objective in Equation \ref{total_object} can be be re-formulated as:\
\begin{gather}
    \mathcal{L}_{kd} = \left\{
\begin{array}{rcl}
      -\sum_{k=1}^{|V|} q(y_k) \notag \cdot \log p(y_k) ,
      {y \in \mathcal{S}_{Hard}} \\
0~~~~~~~~~~~~~~~~, { y  \in \mathcal{S}_{Easy}}
\end{array} \right.
\end{gather}
where we simplify the notation of $p$ and $q$ for clarity.
\paragraph{Global-level Selection (GLS).}  Limited by the number of words in a mini-batch, batch-level selection only reflects the current batch's CE distribution and can not represent the real global CE distribution of the model very well. 
In addition, the batch-level method makes our relative difficulty measure easily affected by each local batch's composition.
The optimal approach to get the global CE distribution is to traverse all training set words and calculate their CE to get the real-time distribution after each model update. 
However, this brings a formidable computational cost and is not realistic in training.

Therefore, as a proxy to optimal way, we extend batch-level selection to global-level selection by dexterously using a First-In-First-Out (FIFO) global queue $\mathcal{Q}$. 
At each training step, we push batch words' CE into FIFO global queue $\mathcal{Q}$ and pop out the `Oldest' words' CE in the queue to retain the queue's size.
Then, we sort all CE values in the queue and calculate the ranking position of each word. 
The storage of queue is much bigger than a mini-batch so that we can evaluate the current batch's CEs with more words, which reduces the fluctuation of CE distribution caused by the batch-level one. Algorithm~1 details the entire procedure.

\begin{algorithm}[t]
\caption{Global-level Selection}
\hspace*{0.02in} {\bf Input:}  
B: mini-batch, $\mathcal{Q}$: FIFO global queue, $\mathcal{T}$: teacher model, $\mathcal{S}$: student model

\begin{algorithmic}[1]
\label{algorithmic}
\For{each $word_i$ in B}
    \State Compute $\mathcal{L}_{ce}$ of $word_i$ by Equation \ref{cross_entorpy}
    \State Compute $\mathcal{L}_{kd}$ of $word_i$ by Equation \ref{kd}
    \State Push $\mathcal{L}_{ce}$ to $\mathcal{Q}$
    \If{${L}_{ce}$ in $top\_{r\%}(\mathcal{Q})$}
        \State $Loss_{i} \gets \mathcal{L}_{ce} + \alpha \cdot \mathcal{L}_{kd}$
    \Else
        \State $Loss_{i} \gets \mathcal{L}_{ce}$
    \EndIf
    \State $Loss \gets Loss + Loss_i $
\EndFor

\State Update  $\mathcal{S}$ with respect to $Loss$
\end{algorithmic}
\end{algorithm}

\section{Experiments}

We carry out experiments on two large-scale machine translation tasks: WMT'14 English-German (En-De) and  WMT'19 Chinese-English (Zh-En).
\subsection{Setup}
\paragraph{Datasets.}
For WMT'14 En-De task, we use 4.5M preprocessed data, which is tokenized and split using byte pair encoded (BPE) \citep{sennrich-etal-2016-neural} with 32K merge operations and a shared vocabulary for English and German. We use \emph{ newstest2013} as the validation set and \emph{ newstest2014} as the test set, which contain 3000 and 3003 sentences, respectively.

For the WMT'19 Zh-En task, we use 20.4M
preprocessed data, which is tokenized and split
using 47K/32K BPE merge operations for source and target languages. We use \emph{ newstest2018} as our validation set and \emph{ newstest2019} as our test set, which contain 3981 and 2000 sentences, respectively.

\paragraph{Evaluation.}
For evaluation, we train all the models with a maximum of 300K steps for WMT En-De'14 and WMT'19 Zh-En. We choose the model which performs the best on the validation set and report its performance on test set. We measure case sensitive BLEU calculated by \emph{ multi-bleu.perl}\footnote{\url{https://github.com/moses-smt/mosesdecoder/blob/master/scripts/generic/multi-bleu.perl}} and  \emph{ mteval-v13a.pl}\footnote{\url{https://github.com/moses-smt/mosesdecoder/blob/master/scripts/generic/mteval-v13a.pl}} with significance test \citep{koehn2004statistical} for WMT'14 En-De and WMT'19 Zh-En, respectively.

\paragraph{Model and Hyper-parameters.} 

\begin{figure}[!t] 
\centering 
\includegraphics[width=0.4\textwidth]{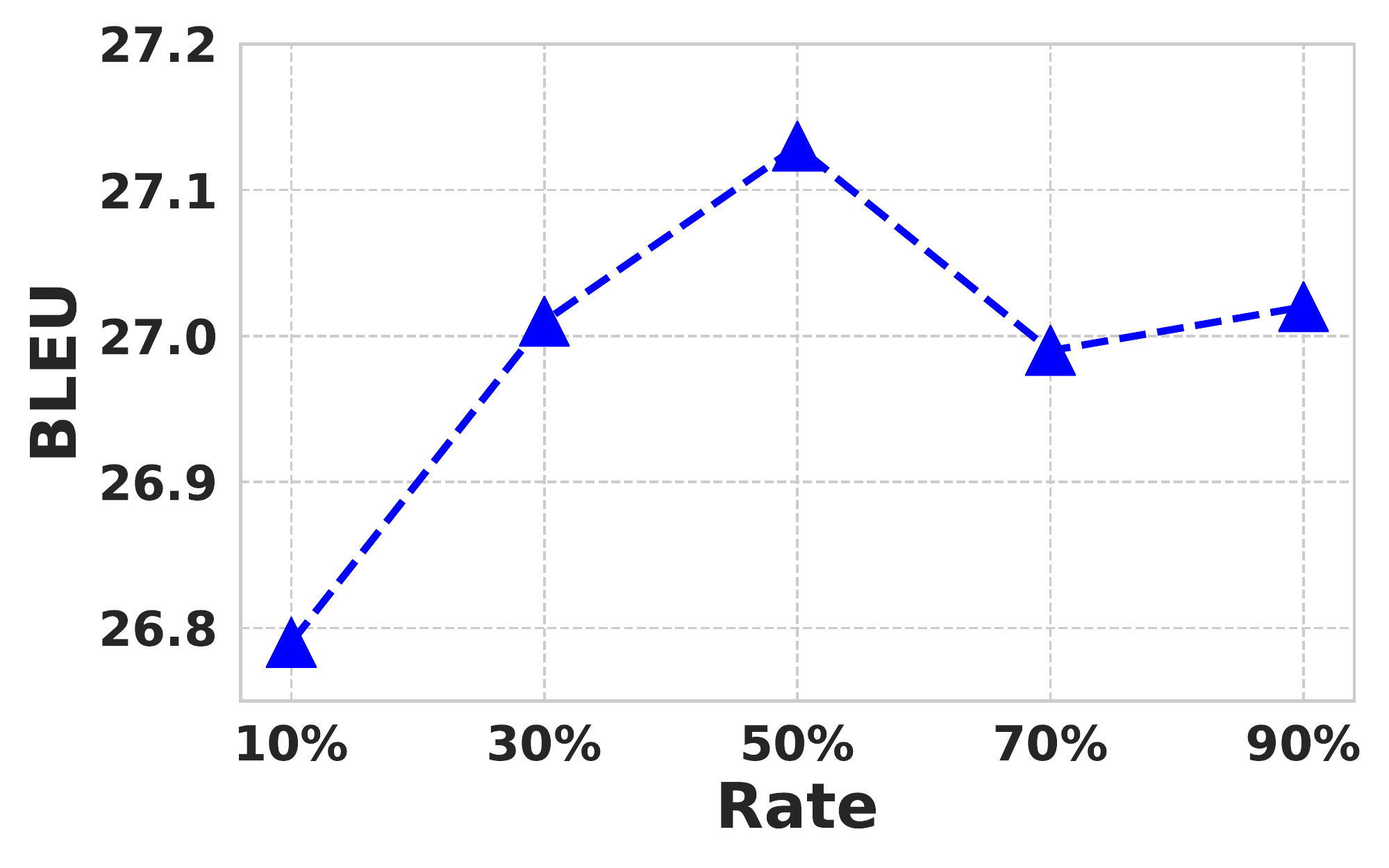} 
\caption{ BLEU score (\%) with different $r\%$ on validation set of WMT’14 En-De.}
\label{Fig.hyper_ende_rate}
\end{figure}

Following the setting in \citet{vaswani2017attention}, we carry out our experiments on standard Transformer \citep{vaswani2017attention} with the fairseq toolkit \cite{ott2019fairseq}. 
By default, we use Transformer (Base), which contains six stacked encoder layers and six stacked decoder layers as both teacher model and student model. 
To verify our approaches can be applied to a stronger teacher and student models, we further use deep Transformers with twelve encoder layers and six decoder layers. 
In training processing, we use Adam optimizer with $\beta_{1} = 0.9$, $\beta_{2} = 0.98$, learning rate is 7e-4 and dropout is 0.1.
All experiments are conducted using 4 NVIDIA P40 GPUs, where the batch size of each GPUs is set to 4096 tokens. And we accumulate the gradient of parameters and update every two steps. The average runtimes are 3 GPU days for all experiments.

There are two hyper-parameters in our experiment, i.e., distil rate $r\%$ and global queue size ${Q}_{size}$. For distil rate $r\%$, the search space is [10\%, 30\%, 50\%, 70\%, 90\%]. The search result of $r\%$ is shown in Figure \ref{Fig.hyper_ende_rate}, we can find that the performance is sensitive to the value of r\%. When the ratio is smaller than 50\%, the increase of ratio is consistent with the BLEU score increases, and the best performance peaks at 50\%. We directly apply the distil rate $r\%$ to the WMT'19 Zh-En task without extra searching.
Besides, We set the $\mathcal{Q}_{size}=$ 30K for WMT'14 En-De.  
For larger dataset WMT'19 Zh-En, we enlarge the $Q_{size}$ to from 30K to 50K and keep word rate unchanged. The hyper-parameter search of $\mathcal{Q}_{size}$ can be found in Section 6.4.

\paragraph{Compared Methods.} We compare our method with several existing NMT systems (KD and others):

\begin{itemize}
\vspace{-5pt}
\setlength{\itemsep}{3pt}
\setlength{\parsep}{0pt}
\setlength{\parskip}{0pt}

\item \textbf{Word-KD \citep{kim2016sequence}.}  Word-KD is a standard method that distills knowledge equally for each word. The detailed description is in Section  \ref{word_kd}.

\item \textbf{Seq-KD \citep{kim2016sequence}.} Sequence-KD uses teacher generated outputs on training corpus as an extra source. The training loss can be formulated as:
   {
    \begin{align} \label{seq_kd} \nonumber
        \mathcal{L}_{seq\_kd}  = - \sum_{j=1}^{J} \sum_{k=1}^{|V|}\mathbbm{1}\{\hat{y}_j = k\} \\ \times \log p(y_j = k| \bm{\hat{y}}_{<j},\bm{x}; \theta)  ,
    \end{align}
   }
where $\bm{\hat{y}}$ denotes the sequence predicted by teacher model from running beam search, $J$ is the length of target sentence.

\item \textbf {Bert-KD \citep{chen2020distilling}.}  This method leverages the pre-trained Bert as teacher model to help NMT model improve machine translation quality. 

\item \textbf {Other Systems.} We also include some existing methods based on Transformer(Base) for comparison, i.e., \citet{zheng2019dynamic, so2019evolved,tay2020synthesizer}.

\end{itemize}

\subsection{Main Results}

\paragraph{Results on  WMT'14 English-German.}

\begin{table}[!t]
\begin{center}
\begin{tabular}{ll|ll}
\hline
\multicolumn{2}{l|} {Models} & \multicolumn{1}{l}{En-De} &  \multicolumn{1}{c}{$\Delta$}   \\ \hline 
\multicolumn{4}{c}{\emph{Existing NMT systems}}    \\ \hline
\multicolumn{2}{l|}{\citet{vaswani2017attention}}   &    \multicolumn{1}{l}{27.30} & \multicolumn{1}{c}{ref}  \\
\multicolumn{2}{l|}{\citet{vaswani2017attention} (Big)}   &    \multicolumn{1}{l}{28.40} & \multicolumn{1}{c}{+1.10}    \\ 
\multicolumn{2}{l|}{\citet{chen2020distilling}}   &    \multicolumn{1}{l}{27.53} & \multicolumn{1}{c}{+0.23}     \\
\multicolumn{2}{l|}{ \citet{zheng2019dynamic} }   &    \multicolumn{1}{l}{28.10} & \multicolumn{1}{c}{+0.80}    \\ 
\multicolumn{2}{l|}{ \citet{so2019evolved} }   &    \multicolumn{1}{l}{28.40} & \multicolumn{1}{c}{+1.10}    \\ 
\multicolumn{2}{l|}{ \citet{tay2020synthesizer} }   &    \multicolumn{1}{l}{28.47} & \multicolumn{1}{c}{+1.17}    \\

\hline 
\multicolumn{4}{c}{\emph{Our Implemented Methods}}    \\ \hline
\multicolumn{2}{l|}{Transformer}   &    \multicolumn{1}{l}{27.29} & \multicolumn{1}{c}{ref}  \\
\multicolumn{2}{l|}{Word-KD} &   \multicolumn{1}{l}{28.14} & \multicolumn{1}{c}{+0.85} \\
\multicolumn{2}{l|}{Seq-KD}   &   \multicolumn{1}{l}{28.15} & \multicolumn{1}{c}{+0.86}  \\

\multicolumn{2}{l|}{Batch-level Selection}  &   \multicolumn{1}{l}{28.42*} & \multicolumn{1}{c}{+1.13} \\ 
\multicolumn{2}{l|}{Global-level Selection}  &   \multicolumn{1}{l}{\textbf{28.57*\dag}}  &  \multicolumn{1}{c}{\textbf{+1.28}}\\ \hline
\end{tabular}
\caption { BLEU scores (\%) on WMT'14 English-German (En-De) task. $\Delta$ shows the improvement compared to Transformer (Base).
`*': significantly ($p < 0.01$) better than Transformer (Base).
`\dag': significantly ($ p < 0.05$) better than the  Word/Seq-KD models. 
}
\label{tab:performance_ende}
\vspace{-15pt}
\end{center}
\end{table}

The results on WMT'14 En-De are shown in Table \ref{tab:performance_ende}. 
In this experiment, both the teacher model and student model are Transformer (Base). 
We also list our implementation of word-level distillation and sequence level distillation \citep{kim2016sequence} method. 

Firstly, compared with the Transformer (Base), our re-implemented word-level and the sequence-level distillation show similar improvements with the BLEU scores up from 27.29 to 28.14 and 28.15, respectively.
Secondly, compared with these already strong baseline methods, our batch-level selective approach further extends the improvement to 28.42, proving the selective strategy's effectiveness. 
Thirdly, our global-level distillation achieves a 28.57 BLEU score and outperforms all previous methods, showing that the better evaluation of words' CE distribution with FIFO global queue helps selection. It is worth noting that our strategy also significantly improves translation quality over all others methods including Word-KD.
Finally, our methods show comparable/better performance than other existing NMT systems and even surpass the Transformer (Big), with much fewer parameters.

\subsection{Analysis}
Even though we find some interesting phenomena and achieve great improvement by selective distillation, the reason behind it is still unclear.  Hence, in this section, we conduct some experiments to analyze and explain the remaining question.

Note that we follow the previous partition and comparison method in this section and divide the samples with/without KD loss defined in our selection strategy as $\mathcal{S}_{Hard}$/$\mathcal{S}_{Easy}$.

\begin{figure}[!t] 
\centering 
\includegraphics[width=0.45\textwidth]{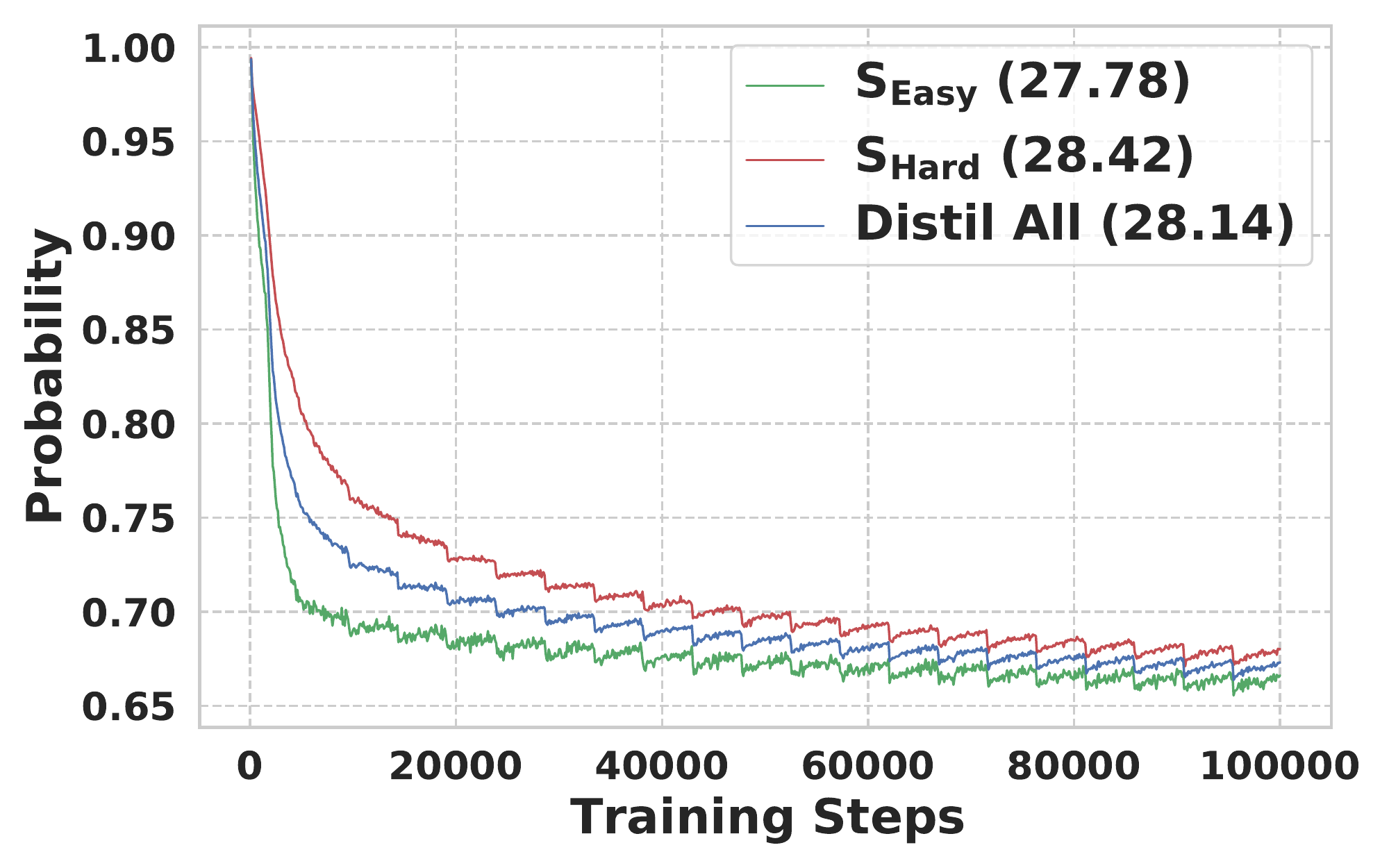} 
\caption{The probability for gradients of $\mathcal{L}_{kd}$ and $\mathcal{L}_{ce}$ pointing the same direction. } 
\label{Fig.cosine_simi}
\end{figure}

\paragraph{Conflict on Different Parts.}
The first question is that why our methods surpass the Word-KD with more knowledge. To answer this question, we collect the statistics on the gradient difference between knowledge distillation loss and cross-entropy loss on the ground-truth label for $\mathcal{S}_{Hard}$ and $\mathcal{S}_{Easy}$.

Here, we study gradients over the output distributions, which are directly related to the model's performance.
Particularly, decoder maps target sentences $\bm{y}=(y_1^*,...,y_m^*)$ to their corresponding hidden representation $\bm{h}=(h_1,...,h_m)$. 
For words in target sequence, the prediction logits $l \in R^{d_{model} \times |V|}$ is given by: 
{
    \begin{gather} \label{distribution}
        l = h^T \mathcal W \\
        p = Softmax(l)
    \end{gather}
}where $h \in R^{d_{model}}$ is the layer output of transformer decoder, $\mathcal{W} \in R^{d_{model} \times |V|}$ is projection matrix.
Then, the gradient respect to $l$ from golden cross-entropy loss can be denotes as $\nabla_{l}{\mathcal{L}_{ce}}$. 
The gradient from distillation loss can be denotes as  
$\nabla_{l}{\mathcal{L}_{kd}}$. 
Next, we calculate the probability that $\nabla_{l}{\mathcal{L}_{ce}}$ and $\nabla_{l}{\mathcal{L}_{kd}}$ share the same direction.

Figure \ref{Fig.cosine_simi} presents the results with the probability that gradients agree with each other during training.
We observe that $\mathcal{S}_{Easy}$ (green line) is consistently lower than distillation with all words (blue line) and $\mathcal{S}_{Hard}$ (red line), which means $\mathcal{S}_{Easy}$ has more inconsistency with ground-truth. 
Combining with the BLEU performances, we argue this consistency leads to the risk of introducing noise and disturbs the direction of parameter updating. 

Besides, the agreement of Distill-All (blue line in Fig) lies in the middle of two halves. It proves that $\mathcal{S}_{Easy}$ and $\mathcal{S}_{Hard}$ compromise with each other on some conflicts. It also proves that there exist some conflicts between the knowledge in $\mathcal{S}_{Easy}$ and $\mathcal{S}_{Hard}$.

\begin{figure}[!t] 
\centering
\includegraphics[width=0.4\textwidth]{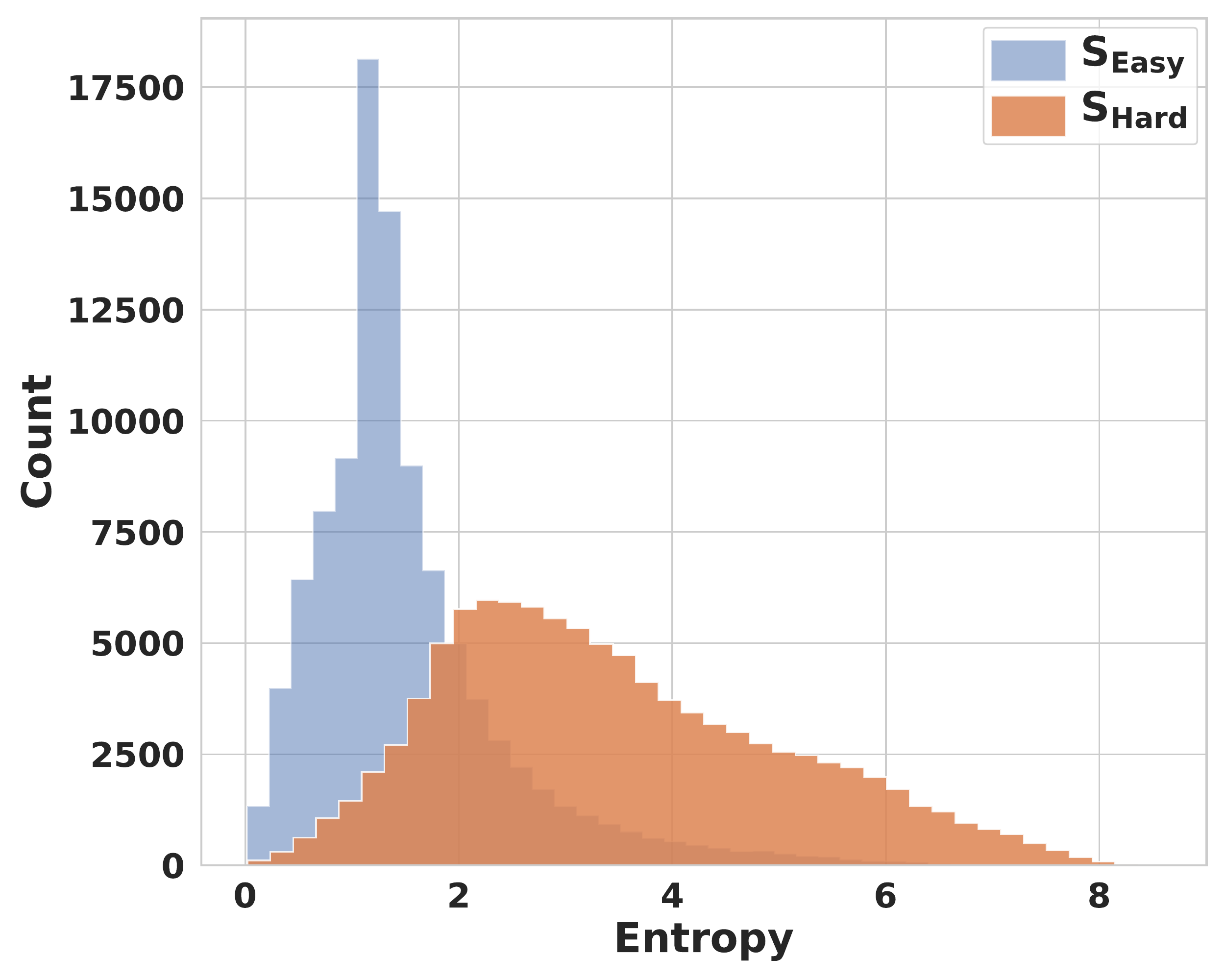} 
\caption{The entropy of prediction distribution of teacher model for different parts.}
\label{Fig.ce_easy_hard}
\vspace{-5pt}
\end{figure}

\paragraph{Knowledge on Different Parts.}
In our approaches, we select the transferring samples from the student model's point of view. However, in previous literature, they commonly consider knowledge from the teacher's perspective. 
Hence, in this section, we study the correlation between these two perspectives.

Because previous studies commonly regard teacher's soft-labels contain dark knowledge \citep{dong2019distillation}, we take the entropy of teacher's prediction as a proxy.
Concretely, we randomly select 100K tokens in the training set and calculate the entropy of distribution predicted by the teacher model for both $\mathcal{S}_{Hard}$ and $\mathcal{S}_{Easy}$. 
As shown in Figure \ref{Fig.ce_easy_hard}, we notice that the $\mathcal{S}_{Easy}$'s entropy distribution is more concentrated in range (0, 4) and peaks around 1.2.
In contrast, the $\mathcal{S}_{Hard}$'s entropy distribution is more spread out. 
The overall distribution shifts to higher entropy, which indicates $\mathcal{S}_{Hard}$ tends to provide a smoother supervision signal.
Consequently, we conclude that even though our selective strategy comes from the student's perspective, it also favors samples with abundant dark knowledge in teacher's perspective.
To some extent, this explains why the $\mathcal{S}_{Hard}$' knowledge benefits distillation performance more.

\subsection{Generalizability}
\paragraph{Results on  WMT'19 Chinese-English.}
\begin{table}[!t]
\begin{center}
\begin{tabular}{ll|ll||ll}
\hline
\multicolumn{2}{l|} {Models} &\multicolumn{1}{c}{Zh-En} &  \multicolumn{1}{c}{$\Delta$} \\ \hline 
\multicolumn{2}{l|}{Transformer (Base)}  &   \multicolumn{1}{l}{25.73} & \multicolumn{1}{c}{ref}    \\ 
  
\multicolumn{2}{l|}{Word-KD} & \multicolumn{1}{l}{26.21} & \multicolumn{1}{c}{+0.48} \\
\multicolumn{2}{l|}{Seq-KD} & \multicolumn{1}{l}{27.27} & \multicolumn{1}{c}{+1.54} \\
\hline

\multicolumn{2}{l|}{Word-KD + Ours} & \multicolumn{1}{l}{26.62*}   &  \multicolumn{1}{c}{+0.89}\\
\multicolumn{2}{l|}{Seq-KD + Ours} & \multicolumn{1}{l}{27.61*}   &  \multicolumn{1}{c}{+1.88}\\
\hline
\end{tabular}
\caption {BLEU scores (\%) on WMT'19 Chinese-English (Zh-En) task. $\Delta$ shows the improvement compared to Transformer (Base). `*': significantly ($ p < 0.01$) better than the  Transformer (Base).
}

\label{tab:performance_zhen}

\end{center}
\end{table}
We also conduct experiments on the larger WMT'19 Zh-en dataset (20.4M sentence pairs) to ensure our methods can provide consistent improvements across different language pairs. 

As shown in Table \ref{tab:performance_zhen}, our method still significantly outperforms the Transformer (Base) with +0.89. Compared with the Word-KD, our approach consistently improves with +0.41 BLEU points. Besides, we also find that Seq-KD with our methods extends the improvement of BLEU score from 27.27 to 27.61. This indicates that our selective strategy is partially orthogonal to the improvement of Seq-KD and maintains generalizability.
In summary, these results suggest that our methods can achieve consistent improvement on different sized datasets across different language pairs. 

\begin{table}[!t]
\begin{center}
\begin{tabular}{ll|ll}
\hline
\multicolumn{2}{l|} {Models} & \multicolumn{1}{c}{En-De} &  \multicolumn{1}{c}{$\Delta$}   \\ \hline 
\multicolumn{2}{l|}{Deep Transformer (12 + 6)}   &    \multicolumn{1}{l}{27.94} & \multicolumn{1}{c}{ref}  \\ 
\multicolumn{2}{l|}{Word-KD} &   \multicolumn{1}{l}{28.90} & \multicolumn{1}{c}{+0.96} \\ \hline

\multicolumn{2}{l|}{Ours}  &   \multicolumn{1}{l}{\textbf{29.12*}}  &  \multicolumn{1}{c}{+1.18}\\ \hline
\end{tabular}
\caption {BLEU scores (\%) on WMT'14 English-German (En-De) task. Here we use Deep Transformers (12 encoders and 6 decoders) for both the teacher and student model.  $\Delta$ shows the improvement compared to Deep Transformer (12 + 6).
`*': significantly ($p < 0.01$) better than Deep Transformer (12 + 6).
}
\label{tab:performance_big_small}

\end{center}
\end{table}

\paragraph{Results with Larger Model Size.}
Here, we investigate how our method is well-generalized to larger models.  
We use a deep transformer model with twelve encoder layers and six decoder layers for our larger model experiments.
As shown in Table \ref{tab:performance_big_small}, Deep Transformer (12 + 6) and Word-KD have already achieved strong performance with up to 28.90 BLEU points, and our method still outperforms these baselines (29.12 BLEU). It proves our methods' generalizability to larger models.

\subsection{Effect of the Global Queue}

This section analyzes how $\mathcal{Q}_{size}$ affects our model's performance.
As mentioned before, $\mathcal{Q}_{size}$ denotes the size of the global FIFO queue, which affects simulating the word cross-entropy distribution of the current model.

\begin{figure}[!t]
\begin{subfigure}{.45\textwidth}
	\centering
	\includegraphics[width=\textwidth]{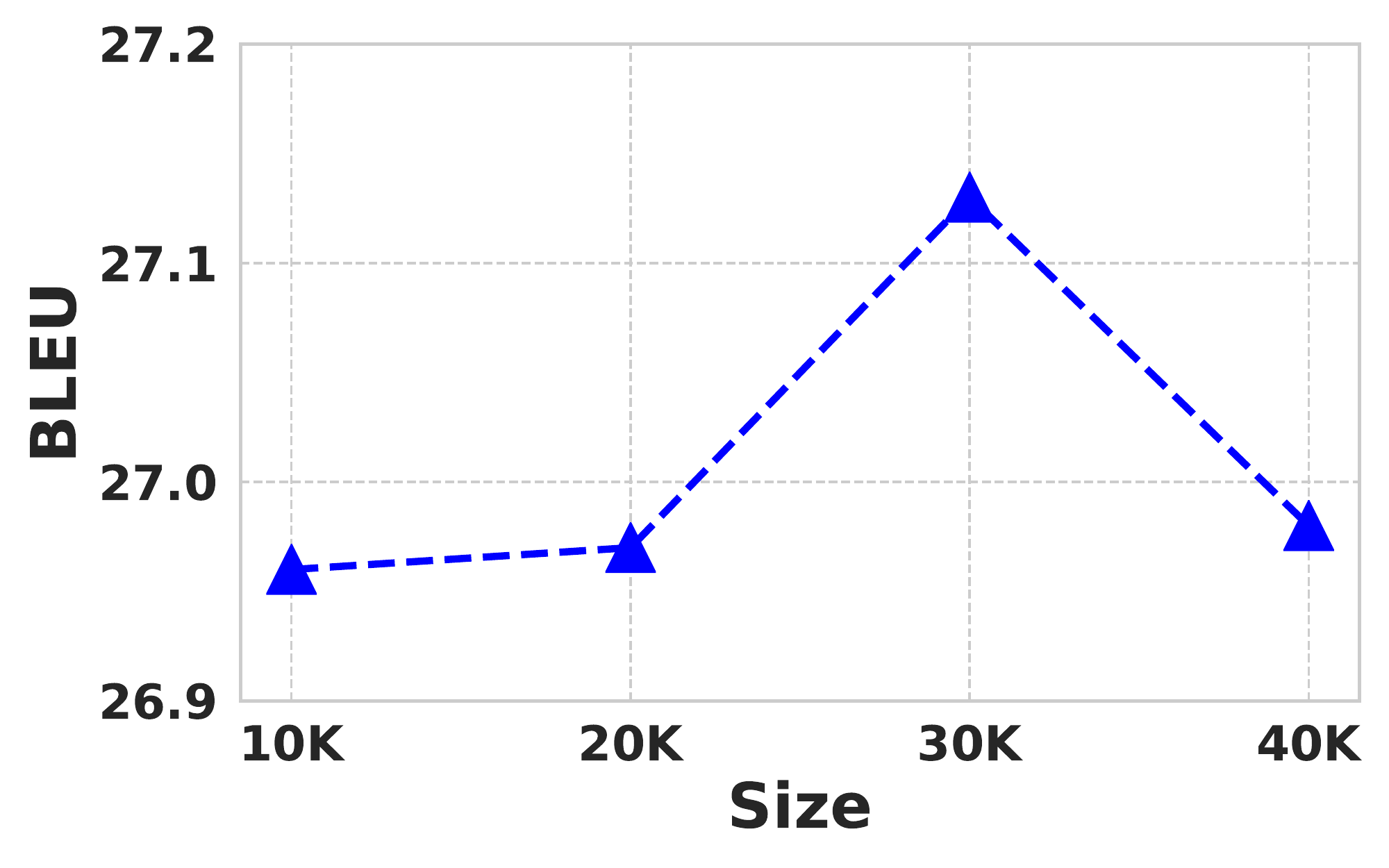}
	\caption{En-De}
\end{subfigure}
\begin{subfigure}{.45\textwidth}
	\includegraphics[width=\textwidth]{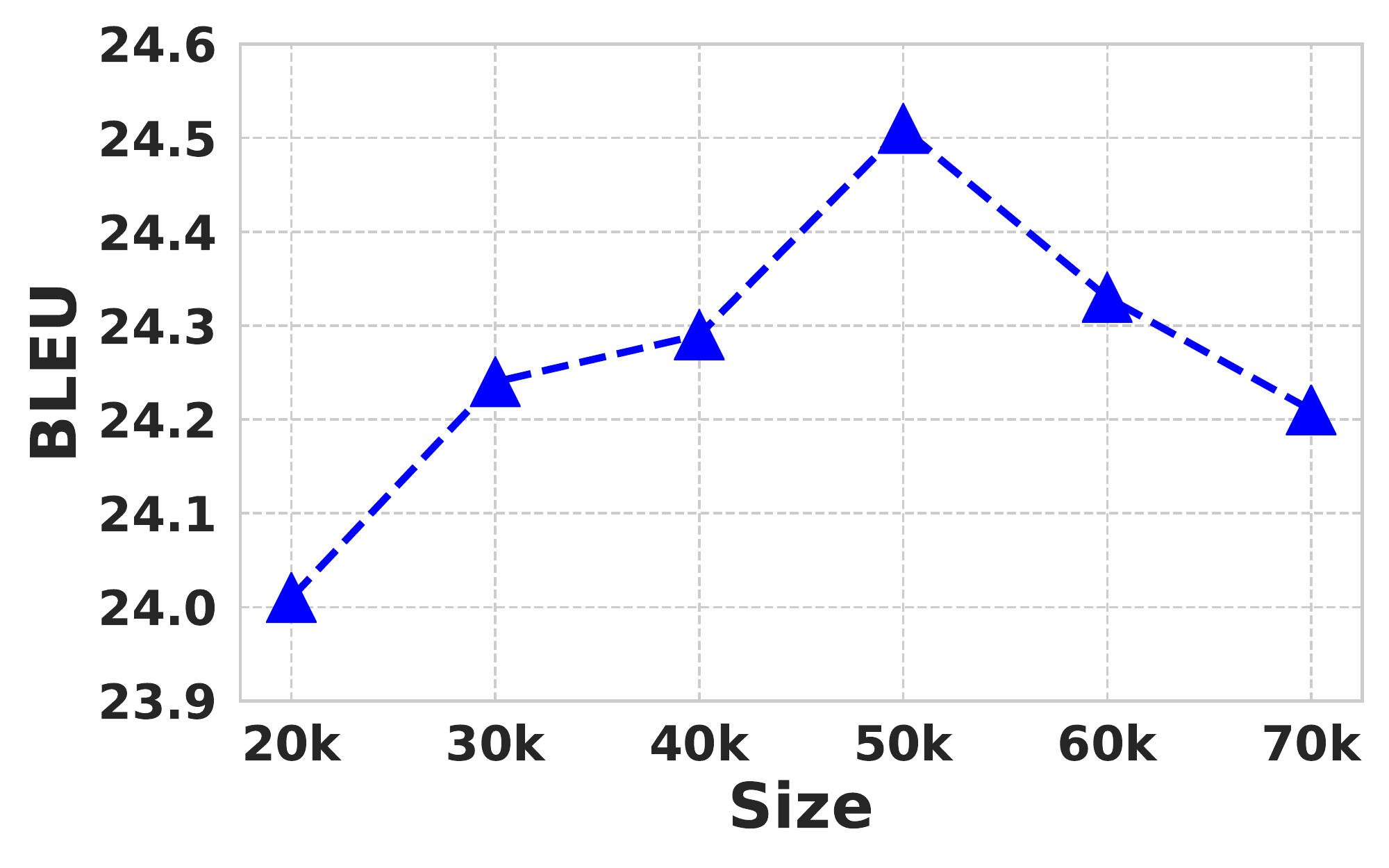}
	\caption{Zh-En}
\end{subfigure}
\caption{ BLEU score (\%) with different $\mathcal{Q}_{size}$ on WMT’14 En-De and WMT'19 Zh-En validation set.} 
\label{Fig.hyper_param_size}
\end{figure}

Figure \ref{Fig.hyper_param_size} shows the search results of $\mathcal{Q}_{size}$. 
We can find that smaller and larger queue size both hurts the BLEU scores. Besides, 30K and 50K of queue size are the best for WMT’14 En-De and WMT’19 Zh-En, respectively. This also accords with our intuition that smaller $\mathcal{Q}_{size}$ degrades the global-level queue to batch level, and larger $\mathcal{Q}_{size}$ slows down the update of CE distribution.

\begin{figure}[!t] 
\centering
\includegraphics[width=0.48\textwidth]{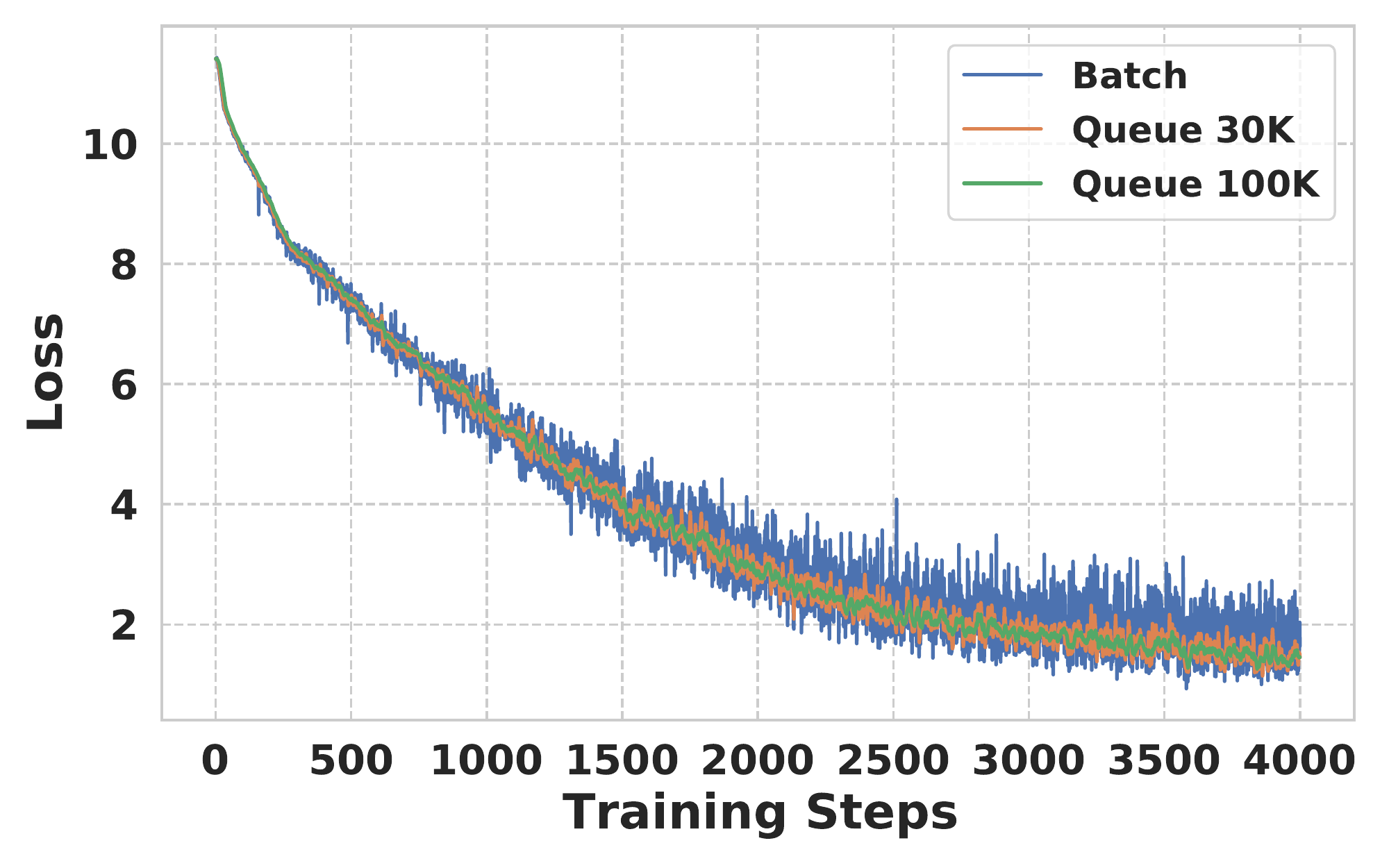} 
\caption{Partition point for $\mathcal{S}_{Hard}$ and $\mathcal{S}_{Easy}$, with respect to different strategies. Batch-level selection clearly suffers from large fluctuations and high variance.} 
\label{Fig:effect_queue_size} 
\end{figure}

Figure \ref{Fig:effect_queue_size} plots the partition Word CE of $\mathcal{S}_{Hard}$ and $\mathcal{S}_{Easy}$ for batch-level and global-level selection. We can see that, as the training progresses, batch-level selection starts to suffer from the high variance because of each batch's randomness. 
Selections with FIFO queue drastically reduce the variance and make a reasonable estimation of global CE distribution.
These findings prove the effectiveness of our proposed FIFO queue.

\section{Conclusion}
In this work, we conduct an extensive study to analyze the impact of different words/sentences as the carrier in knowledge distillation. 
Analytic results show that distillation benefits have a substantial margin, and these benefits may not collaborate with their complementary parts and even hurt the performance. 
To address this problem, we propose two simple yet effective  strategies, namely the batch-level selection and global-level selection. 
The experiment results show that our approaches can achieve consistent improvements on different sized datasets across different language pairs.

\section*{Acknowledgments}
We would like to thank the anonymous reviewers for their valuable comments and suggestions to improve this paper.

\bibliographystyle{acl_natbib}
\bibliography{anthology,acl2021}
 
\end{document}